\documentclass[a4paper, 10pt, conference]{ieeeconf}      

\IEEEoverridecommandlockouts                              

\usepackage{url}
\usepackage{comment}
\usepackage{amsmath,amssymb,amsfonts, mathtools}
\usepackage{graphicx}
\usepackage{bm}
\usepackage{epstopdf}
\usepackage{placeins}
\usepackage{bm}
\usepackage{cite}
\let\labelindent\relax
\usepackage[inline]{enumitem}
\usepackage{multicol}
\usepackage{multirow}
\usepackage{tikz}
\usepackage{caption}
\usepackage{subcaption}
\usepackage{tabto}
\usepackage{siunitx}
\usepackage{gensymb}
\usepackage{supertabular}
\usepackage{booktabs}

\title{\LARGE \bf
Design of Fuzzy Logic Parameter Tuners for Upper-Limb Assistive Robots
}

\author{Christopher Coco Jr., Jonathan Spanos, Hamid Osooli, and  Reza Azadeh
\thanks{Authors are with the Persistent Autonomy and Robot Learning (PeARL) Lab, University of Massachusetts Lowell, Lowell, MA 01854, USA {\tt\small \{christopher\_coco, jonathan\_spanos, hamid\_osooli\}@student.uml.edu}, \tt\small reza@cs.uml.edu}
}

\begin{document}
\bstctlcite{IEEEexample:BSTcontrol}

\maketitle
\thispagestyle{empty}
\pagestyle{empty}

\begin{abstract}

Assistive Exoskeleton Robots are helping restore functions to people suffering from underlying medical conditions. These robots require precise tuning of hyper-parameters to feel natural to the user. The device hyper-parameters often need to be re-tuned from task to task, which can be tedious and require expert knowledge. To address this issue, we develop a set of fuzzy logic controllers that can dynamically tune robot gain parameters to adapt its sensitivity to the user's intention determined from muscle activation. The designed fuzzy controllers benefit from a set of expert-defined rules and do not rely on extensive amounts of training data. We evaluate the designed controllers with three different tasks and compare our results against the manually tuned system. Our preliminary results show that our controllers reduce the amount of fighting between the device and the human, measured using a set of pressure sensors.



\end{abstract}

\section{Introduction}

Assistive robotics is a growing field that aims to help people in need. Among others, exoskeleton robots and assistive orthoses can help restore some daily actions (e.g., picking objects or opening doors) that may be difficult or impossible due to underlying medical conditions~\cite{shoemaker2018myoelectric, chang2023myoelectric}. Although these robots are usually equipped with effective low-level controllers (e.g., PID controllers), the high-level parameters of the device (and sometimes the low-level control parameters such as the PID gains) need to be re-tuned manually to feel natural to the user. The device parameter tuning process can be done by experts. However, training the user or their caregivers to perform the parameter tuning for different tasks under different circumstances is a challenging process and could be tedious. As a result, researchers have been studying methods for hyper-parameter tuning using high-level controllers. A common theme in this area is the employment of the data-driven control methods. For instance, several reinforcement learning-based methods have been developed to learn sEMG-based control policies for upper-limb exoskeletons ~\cite{hamaya2016learning,hamaya2017learning,deisenroth2013gaussian}. Several other methods rely on the advantages of neural networks~\cite{cao2006neural, medina2021control}. Both categories include algorithms that require extensive amounts of data for performing system identification and training.

Our study, on the other hand, seeks to integrate the benefits of data-driven methods with the robustness of rule-based methods~\cite{weiss1995rule}. To achieve this goal, we propose a fuzzy logic system~\cite{wang1996course, ahmadzadeh2005modeling, ahmadzadeh2013autonomous} that can utilize the expert knowledge in the form of a set of rules designed by a human expert. For this study, we focus on the parameter tuning of upper-limb assistive robots and consider the MyoPro 2 device (Myomo Inc., Cambridge, MA) in our experiments. This device is an elbow and hand exoskeleton designed to help restoring functionality to the wearer’s paralyzed or weakened upper body. To enhance user experience, the hyper-parameters of the robot, two gains in this case, must be tuned properly. We design two fuzzy logic controllers, one for each gain, to enable real-time parameter tuning. The ultimate goal of this research is to enable the robot to adapt from task to task under different conditions, improving the quality of human assistance experience. In this paper, we identify the hyper-parameters and critical factors affecting them. We then explain our controller design process and experimental setup. We evaluate the designed controllers with three tasks under several different conditions and compare the results against scenarios where the robot was tuned manually. Our preliminary results show the effectiveness of the proposed controllers for the hyper-parameter tuning.

\begin{figure}
    \centering
    \includegraphics[width=0.9\columnwidth]{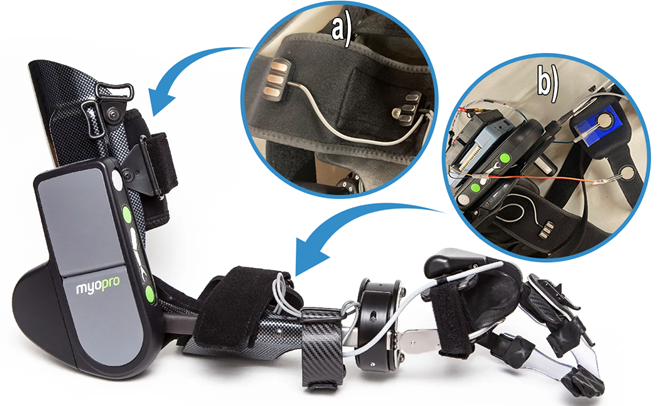}
    \caption{\small{MyoPro 2 orthosis used in our research. (a) integrated sEMG sensors and (b) integrated pressure sensors for the arm joint.}}
    \label{fig:device}
\end{figure}

\section{Device Overview}

The MyoPro 2 orthosis (Fig.~\ref{fig:device}) is an upper-limb exoskeleton designed to help with everyday activities. The device has two motor-controlled degrees of freedom (DoF). The elbow joint governs arm movements (i.e., extension and flexion) while the wrist joint enables hand movements (i.e., open and close). The device reads and interprets the muscle activation patterns into motor commands using two surface electromyography (sEMG) sensors. In this paper, we only focus on the arm movements which are continuous.



The MyoPro 2 robot has four different control modes: (a) \textit{Standby mode} in which both of the motors do not receive readings from the sEMG sensors, (b) \textit{Bicep mode} where the device only receives information from the bicep sensor which allows only for flexion movement, (c) \textit{Tricep mode} where the device only receives information from the tricep sensor which only allows for extension movement, and (d) \textit{Dual mode} which gets information from both sensors which allows for extension/flexion movements. In this paper, we focus on the \textit{dual mode} which is the mode applicable to most everyday tasks. Although the dual mode can be set with three different movement types: constant, proportional, and exponential, our research focuses on the proportional mode that allows the device to adjust the speed proportionally. In other words, the device activates by measuring the changes in the dominant and opposing effort ($E_d$ and $E_o$) weighted proportionally by a gain value ($k_p$) as 
\begin{equation}
    S = k_p \Delta E,
\end{equation}
\noindent where $S$ is the joint speed, $\Delta E=E_d - E_o$, and effort is defined as the ratio between the current and maximum values of the sEMG signal. The dominant and opposing efforts refer to the maximum and minimum value between the bicep and the tricep efforts, respectively. Whenever the  effort difference $\Delta E$ surpasses a preset threshold for the muscle of the dominant effort the device then moves in that direction. Besides choosing a movement type and tuning the bicep/tricep effort thresholds, the user can also adjust the bicep/tricep gains. These gains are used to amplify the sEMG signal. In this paper, we consider tuning the gains instead of the thresholds. The reason is that the gains have a more direct effect on the sensitivity of the sEMG sensors, while the thresholds change upon muscle activation. 

\section{Fuzzy Controller Design}

\begin{figure}
\centering
\includegraphics[width=\columnwidth]{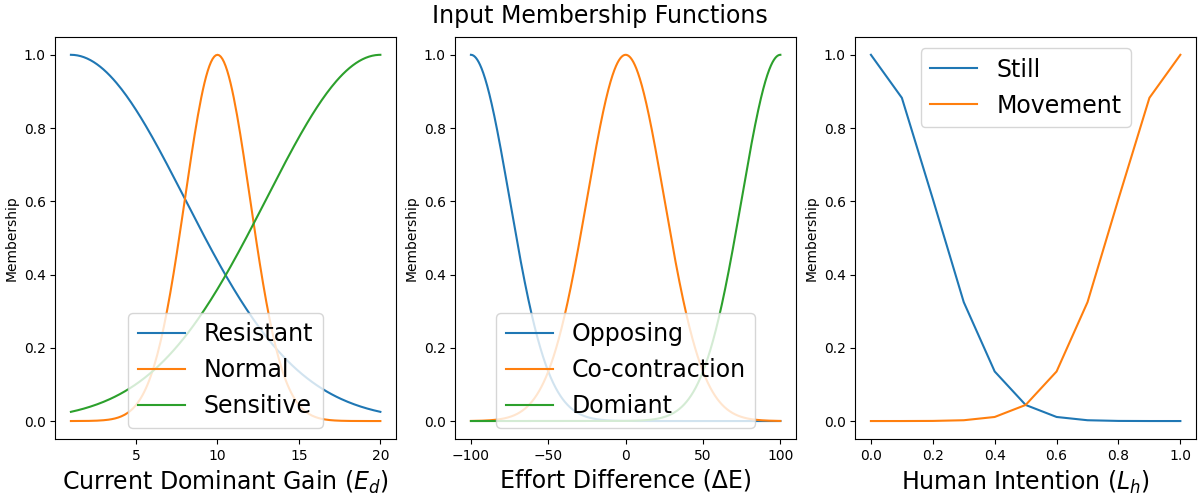}
    \caption{\small{Designed membership functions for the two controllers.}}
    \label{fig:mfs}
\end{figure}

\begin{figure}[b]
\centering
    \includegraphics[width=\columnwidth]{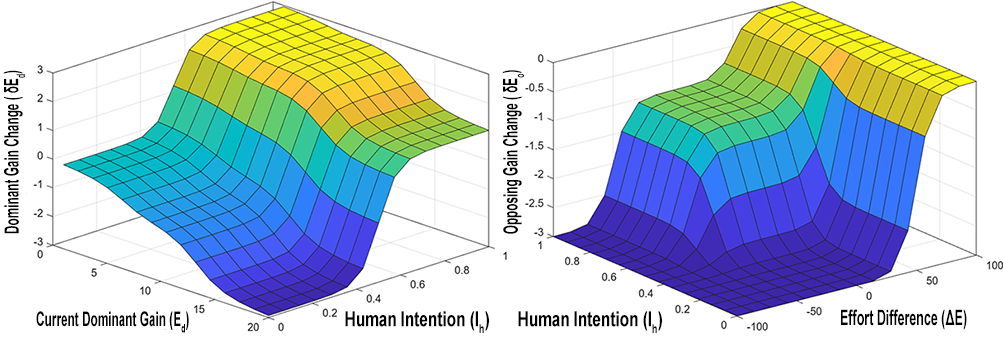}
    \caption{\small{Control surfaces for the controllers: (left) $f_{E_d}$, (right) $f_{E_o}$.}}
    \label{fig:surfaces}
\end{figure}

The overall parameter tuning process of the device depends on three main factors: (a) human intention, (b) current dominant gain $E_d$, and (c) effort difference $\Delta E$. To dynamically tune the parameters of the MyoPro robot, we designed two controllers for adjusting the dominant gain, $\delta E_d$ and the changes in the opposing gain, $\delta E_o$. For both controllers, we used a Takagi-Sugeno-Kang (TSK) fuzzy system~\cite{takagi1985fuzzy} comprising the product inference engine, singleton fuzzifier, and the center average defuzzifier~\cite{wang1996course}. The controller that deals with the dominant gain takes two inputs: the human intention $I_h$ and the current dominant gain $E_d$. We defined two Gaussian Membership Functions (MFs) to represent the human intention. The \textit{still} MF describes the human intention to stay still, whereas the \textit{movement} MF describes the intention to move. The second input (the current $E_d$) is in range $[1, 20]$ . For this input, we define three Gaussian MFs: \textit{resistant}, \textit{normal}, \textit{sensitive}. The output of this controller is a continuous value in range $[-3, 3]$ defined using five constant MFs: \textit{decrease}, \textit{small decrease}, \textit{no change}, \textit{small increase}, and \textit{increase}. This controller can be defined as
\begin{equation}
    \delta E_d = f_{E_d} (I_h, E_d).
\end{equation}

We construct a rule-base for this controller including six IF-ELSE rules. This rule-base, as reported in Table~\ref{table:controller_1}, is continuous, complete, and consistent~\cite{wang1996course}.

\begin{table}[h]
\centering
\caption{\small{Fuzzy rule-base for  dominant gain controller, $f_{E_d}$.}}
\label{table:controller_1}
\addtolength{\tabcolsep}{-0.4em}
\begin{tabular}{l|c|ccc}\toprule
& & \multicolumn{3}{c}{Current Dominant Gain $E_d$}\\\cmidrule(lr){3-5}
 & & Resistant & Normal & Sensitive\\
\midrule
Human
& Movement & Increase & Increase & Small Increase\\
 Intention $I_h$ & Still & No Change & Small Decrease & Decrease\\
\bottomrule
\end{tabular}
\end{table}

The second controller that deals with the opposing gain also takes two inputs: the human intention $I_h$ and the effort difference $\Delta E$. The human intention input in this controller was defined similar to the one in the previous controller. We define the effort difference input, in range $[-100, 100]$, using three Gaussian MFs: \textit{opposing}, \textit{co-contraction}, and \textit{dominant}. Similarly, the output of this controller is also a continuous value, in range $[-3, 3]$, defined using five constant MFs: \textit{decrease}, \textit{small decrease}, \textit{no change}, \textit{small increase}, and \textit{increase}. This controller can be defined as 
\begin{equation}
    \delta E_o = f_{E_o}(I_h, \Delta E).    
\end{equation}
    
We construct a rule-base for this controller including six IF-ELSE rules. This rule-base, as reported in Table~\ref{table:controller_2}, is also continuous, complete, and consistent~\cite{wang1996course}. The designed MFs and the corresponding control surfaces can be seen in Fig.~\ref{fig:mfs} and Fig.~\ref{fig:surfaces}, respectively.

\begin{table}[h]
\centering
\caption{\small{Fuzzy rule-base for  opposing gain controller, $f_{E_o}$.}}
\label{table:controller_2}
\addtolength{\tabcolsep}{-0.4em}
\begin{tabular}{l|c|ccc}\toprule
& & \multicolumn{3}{c}{Effort difference $\Delta E$}\\\cmidrule(lr){3-5}
 & & Dominant & Co-contraction & Opposing\\
\midrule
Human
& Movement & No Change & Small Decrease & Decrease\\
 Intention $I_h$ & Still & No Change & Decrease & Decrease\\
\bottomrule
\end{tabular}
\end{table}

\section{Experiments}

\subsection{Experimental Setup}

We evaluated the designed controllers using a MyoPro 2 orthosis in nine experiments. We compared the performance of the controllers against the manually-tuned device. To evaluate the effectiveness of the controllers, we defined a metric that  measures the amount of \textit{fighting} between the device and the human measured using two pressure sensors. It should be noted that the integration of pressure and sEMG sensors is common in assistive human-robot interaction scenarios~\cite{siu2018implementation}. During the manual tuning, the human used the built-in GUI (called MyConfig) to tune the gains in range $[0.2, 20]$. The default value for each gain was set to 10.

\subsection{Task Design}

\begin{figure}
    \centering
    \includegraphics[width=\columnwidth]{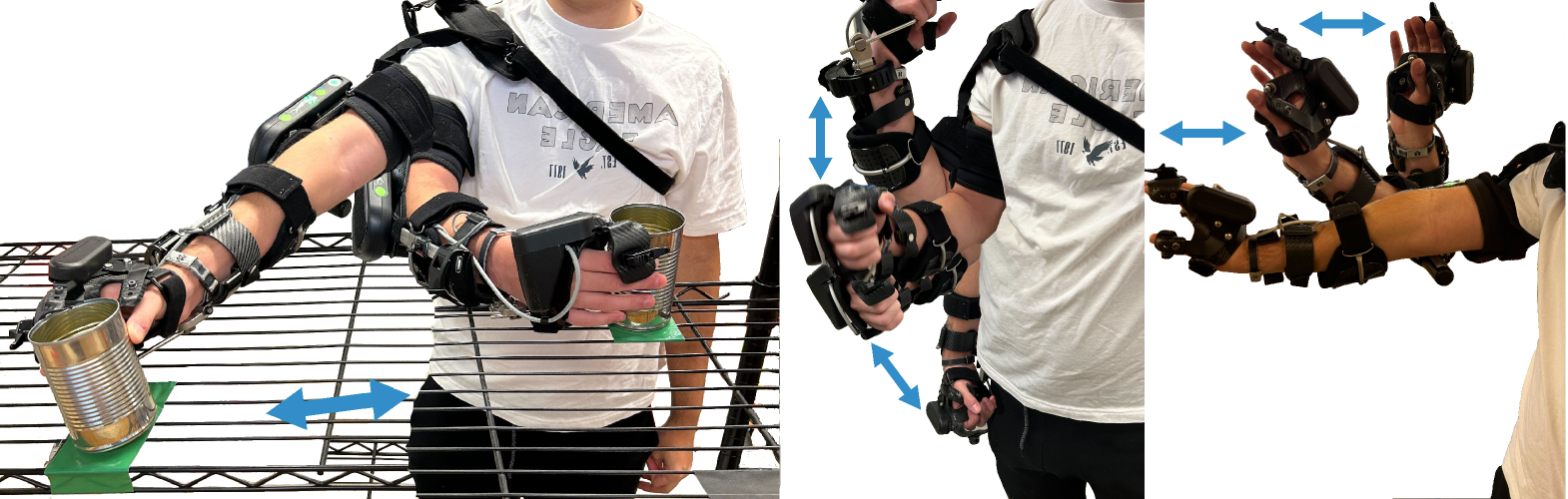}
    \caption{\small{Three tasks designed for evaluation. From left to right: horizontal motion, vertical motion, and pushing motion.}}
    \label{fig:tasks}
\end{figure}

We designed three tasks to include movements with different varieties. The first task, shown in Fig.~\ref{fig:tasks} (middle), is a  vertical motion that resembles a curling exercise. The second task, shown in Fig.~\ref{fig:tasks} (left), is a horizontal motion that includes relocating an empty can. The third task, shown in Fig.~\ref{fig:tasks} (right), is a pushing motion. To collect data, each task was performed three times. Each time we manually set the effort thresholds to a different value to generate different scenarios. The thresholds were set to 10, 20, and 40, representing the \textit{sensitive}, \textit{normal}, and \textit{resistant} scenarios, respectively. For the manual tuning experiments, both gains were set to 10 which is considered the default value for the device. In each test, we collected data for 30 sec. For each scenario, this procedure allowed the user to repeat the vertical, horizontal, and the pushing movements, 5, 6, and 9 times, respectively. After each test, the user rested for a few seconds to reduce the effect of fatigue.

\begin{figure*}[t]
    \centering
    \includegraphics[width=\textwidth]{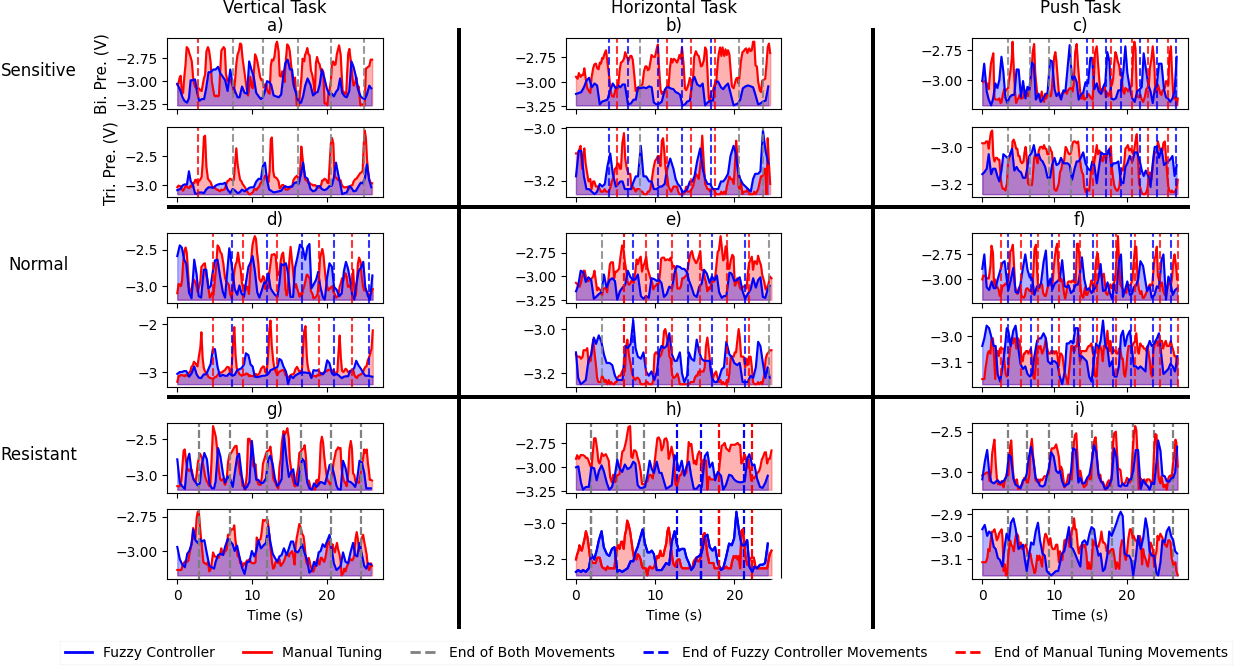}
    \caption{\small{Recorded pressure sensors signals for the sensitive (top row), normal (middle row), and resistant scenarios (bottom row). The plots correspond to the  vertical (left column), horizontal (middle column), and pushing tasks (right column). In each task, the joint angle trajectory is plotted. The vertical dashed lines indicate when one trial of each task was completed and the next one started.}}
    \label{fig:results_res}
\end{figure*}

\subsection{Data Collection \& Pressure Sensors}

To measure the amount of fighting between the user and the device, two MD30-60 pressure sensors were integrated to the MyoPro 2 orthosis. The placement of the sensors can be seen in Fig.~\ref{fig:device}. These sensors were aligned to make contact with the radius and ulna bones. It is important to note that the sEMG sensors can be used for detecting anticipatory signals for grasp and release despite sensitivity to sensor placement~\cite{beckers2015anticipatory, siu2016classification}. The sensor placement, however, is a challenge when using pressure sensors. To mitigate this issue, we incorporate a custom 3D-printed sensor holder, ensuring stable positioning. The printed piece was tied down using a string to keep it in a fixed position. The pressure sensors were powered by two 1.5V AA batteries. To monitor the contact with these sensors the voltage is passed into a BTH-1208LS Data Acquisition System (DAQ). In our setup, the voltage increases when the user fighting with the device increases.

\subsection{Results}

To evaluate the effectiveness of the designed controller compared to the manual tuning case, we recorded the pressure during each task (i.e., vertical, horizontal, push) over different scenarios (i.e., sensitive, normal, resistant). The recorded data can be seen in Fig.~\ref{fig:results_res} where the blue curves indicate the measured signal for when using the fuzzy controllers, while the red curves are for the manually tuned cases. For each curve, we calculated the area under the curve resulting in $A_f$ and $A_m$ (area for the fuzzy controller signal and area for manual tuning). For comparing the signals, we calculated the ratio $r = \frac{\Delta A}{max([A_f, A_m])}$. Results are reported in Table~\ref{table:ratio_b} and Table~\ref{table:ratio_t}. 
Overall, our results show that the designed controllers helped to improve the vertical and the pushing task. It also helped to improve the horizontal task for the bicep but not for the tricep. Looking closer, we noticed the followings:

\textit{1) Vertical Task:} The vertical task saw the most improvement with the controller because the difference in pressure was above 60\% for both sensitive and resistant for the bicep and sensitive and normal for the tricep.

\textit{2) Horizontal Task:} The horizontal task saw improvement for the bicep however, it saw worst results for the tricep. This is most likely due to the positioning of device during the tests. We plan to investigate this issue in the future.

\textit{3) Push Task:} For this task, we noticed significant improvement in the bicep and decent improvement in the tricep.

\begin{table}[h]
\centering
\caption{\small{Calculated area ratio for bicep pressure}}
\label{table:ratio_b}
\addtolength{\tabcolsep}{-0.4em}
\begin{tabular}{ccccc}\toprule
& \multicolumn{4}{c}{Tasks}\\\cmidrule(lr){3-5}
& & Vertical & Horizontal & Pushing \\\midrule
& Sensitive & 65.31\% & 83\% & 40.1\% \\
Scenarios & Normal & 44.4\% & 74\% & 51\% \\
& Resistant & 63\% & 69\% & 47\% \\
\bottomrule
\end{tabular}
\end{table}
\begin{table}[h]
\centering
\caption{\small{Calculated area ratio for tricep pressure}}
\label{table:ratio_t}
\addtolength{\tabcolsep}{-0.4em}
\begin{tabular}{ccccc}\toprule
& \multicolumn{4}{c}{Tasks}\\\cmidrule(lr){3-5}
& & Vertical & Horizontal & Pushing \\\midrule
& Sensitive & 60\% & -38\% & 20\% \\
Scenarios & Normal & 61\% & 0.02\% & 43.3\% \\
& Resistant & 38\% & -34.2\% & 38\% \\
\bottomrule
\end{tabular}
\end{table}



\section{Conclusions \& Future Work}

We proposed a hyper-parameter tuning method for assistive robots using fuzzy logic and designed two fuzzy controllers to dynamically change the sensitivity of the MyoPro 2 orthosis. We evaluated these controllers across three tasks in various scenarios, comparing their effectiveness to manual tuning by measuring user-device fighting during movement. Preliminary results indicate a positive performance enhancement, particularly for vertical and push tasks. However, no significant improvement was observed for the horizontal task involving the tricep. Further investigations will address controller shortcomings through additional experiments incorporating varied movement features and data collection from multiple human subjects.

\addtolength{\textheight}{-12cm}   




\section*{Acknowledgment}

This work was supported by the National Science Foundation (CMMI-2110214). 
\typeout{}
\bibliographystyle{IEEEtran}
\bibliography{bibtex/bib/references}

\begin{thebibliography}{10}
\providecommand{\url}[1]{#1}
\csname url@samestyle\endcsname
\providecommand{\newblock}{\relax}
\providecommand{\bibinfo}[2]{#2}
\providecommand{\BIBentrySTDinterwordspacing}{\spaceskip=0pt\relax}
\providecommand{\BIBentryALTinterwordstretchfactor}{4}
\providecommand{\BIBentryALTinterwordspacing}{\spaceskip=\fontdimen2\font plus
\BIBentryALTinterwordstretchfactor\fontdimen3\font minus \fontdimen4\font\relax}
\providecommand{\BIBforeignlanguage}[2]{{%
\expandafter\ifx\csname l@#1\endcsname\relax
\typeout{** WARNING: IEEEtran.bst: No hyphenation pattern has been}%
\typeout{** loaded for the language `#1'. Using the pattern for}%
\typeout{** the default language instead.}%
\else
\language=\csname l@#1\endcsname
\fi
#2}}
\providecommand{\BIBdecl}{\relax}
\BIBdecl

\bibitem{shoemaker2018myoelectric}
E.~Shoemaker, ``Myoelectric elbow-wrist-hand orthosis with active grasp for patients with stroke: a case series,'' \emph{Canadian Prosthetics \& Orthotics Journal}, vol.~1, no.~2, 2018.

\bibitem{chang2023myoelectric}
S.~R. Chang, N.~Hofland, Z.~Chen, C.~Tatsuoka, L.~G. Richards, M.~Bruestle, H.~Kovelman, and J.~Naft, ``Myoelectric arm orthosis assists functional activities: A 3-month home use outcome report,'' \emph{Archives of Rehabilitation Research and Clinical Translation}, vol.~5, no.~3, p. 100279, 2023.

\bibitem{hamaya2016learning}
M.~Hamaya, T.~Matsubara, T.~Noda, T.~Teramae, and J.~Morimoto, ``Learning assistive strategies from a few user-robot interactions: Model-based reinforcement learning approach,'' in \emph{IEEE International Conference on Robotics and Automation (ICRA)}, 2016, pp. 3346--3351.

\bibitem{hamaya2017learning}
M.~Hamaya, T.~Matsubara, T.~Noda, T.~Teramae, and J.~Morimoto, ``Learning assistive strategies for exoskeleton robots from user-robot physical interaction,'' \emph{Pattern Recognition Letters}, vol.~99, pp. 67--76, 2017.

\bibitem{deisenroth2013gaussian}
M.~P. Deisenroth, D.~Fox, and C.~E. Rasmussen, ``Gaussian processes for data-efficient learning in robotics and control,'' \emph{IEEE transactions on pattern analysis and machine intelligence}, vol.~37, no.~2, pp. 408--423, 2013.

\bibitem{cao2006neural}
H.~Cao, Y.~Yin, D.~Du, L.~Lin, W.~Gu, and Z.~Yang, ``Neural-network inverse dynamic online learning control on physical exoskeleton,'' in \emph{Neural Information Processing: 13th International Conference, ICONIP 2006, Hong Kong, China, October 3-6, 2006. Proceedings, Part III 13}.\hskip 1em plus 0.5em minus 0.4em\relax Springer, 2006, pp. 702--710.

\bibitem{medina2021control}
F.~Medina, K.~Perez, D.~Cruz-Ortiz, M.~Ballesteros, and I.~Chairez, ``Control of a hybrid upper-limb orthosis device based on a data-driven artificial neural network classifier of electromyography signals,'' \emph{Biomedical Signal Processing and Control}, vol.~68, p. 102624, 2021.

\bibitem{weiss1995rule}
S.~M. Weiss and N.~Indurkhya, ``Rule-based machine learning methods for functional prediction,'' \emph{Journal of Artificial Intelligence Research}, vol.~3, pp. 383--403, 1995.

\bibitem{wang1996course}
L.-X. Wang, \emph{A course in fuzzy systems and control}.\hskip 1em plus 0.5em minus 0.4em\relax Prentice-Hall, Inc., 1996.

\bibitem{ahmadzadeh2005modeling}
S.~R. Ahmadzadeh, ``Modeling of hyper-redundant manipulators dynamics and design of fuzzy controller for the system,'' in \emph{International Conference on Integration of Knowledge Intensive Multi-Agent Systems}.\hskip 1em plus 0.5em minus 0.4em\relax IEEE, 2005, pp. 248--253.

\bibitem{ahmadzadeh2013autonomous}
S.~R. Ahmadzadeh, P.~Kormushev, and D.~G. Caldwell, ``Autonomous robotic valve turning: A hierarchical learning approach,'' in \emph{IEEE International Conference on Robotics and Automation}, 2013, pp. 4629--4634.

\bibitem{takagi1985fuzzy}
T.~Takagi and M.~Sugeno, ``Fuzzy identification of systems and its applications to modeling and control,'' \emph{transactions on systems, man, and cybernetics}, no.~1, pp. 116--132, 1985.

\bibitem{siu2018implementation}
H.~C. Siu, A.~M. Arenas, T.~Sun, and L.~A. Stirling, ``Implementation of a surface electromyography-based upper extremity exoskeleton controller using learning from demonstration,'' \emph{Sensors}, vol.~18, no.~2, p. 467, 2018.

\bibitem{beckers2015anticipatory}
N.~Beckers, R.~Fineman, and L.~Stirling, ``Anticipatory signals in kinematics and muscle activity during functional grasp and release,'' in \emph{IEEE 12th International Conference on Wearable and Implantable Body Sensor Networks (BSN)}, 2015, pp. 1--6.

\bibitem{siu2016classification}
H.~C. Siu, J.~A. Shah, and L.~A. Stirling, ``Classification of anticipatory signals for grasp and release from surface electromyography,'' \emph{Sensors}, vol.~16, no.~11, p. 1782, 2016.

\end{thebibliography}




\end{document}